\documentclass[10pt,twocolumn,letterpaper]{article}

\usepackage{cvpr}              

\usepackage{graphicx}
\usepackage{amsmath}
\usepackage{amssymb}
\usepackage{booktabs}
\usepackage{subcaption}
\usepackage{array}
\usepackage{multirow}
\newcommand{\norm}[1]{\left \lVert #1 \right \rVert}

\definecolor{cvprblue}{rgb}{0.21,0.49,0.74}
\usepackage[pagebackref,breaklinks,colorlinks,allcolors=cvprblue]{hyperref}

\def\paperID{12487} 
\def\confName{CVPR}
\def\confYear{2025}

\begin{document}
\title{GalaxyEdit: Large-Scale Image Editing Dataset with Enhanced Diffusion Adapter}


\author{
    Aniruddha Bala$^{1,*}$ \quad Rohan Jaiswal$^{1,*}$ \quad Siddharth Roheda$^1$ \quad Rohit Chowdhury$^1$ \quad Loay Rashid$^2$ \\
    $^1$Samsung R\&D Institute India, Bangalore \qquad $^2$University of California, San Diego \\
    {\tt\small \{aniruddha.b, r.jaiswal, sid.roheda, rohit.c\}@samsung.com, lrashid@ucsd.edu}
}


\maketitle
\def\thefootnote{*}\footnotetext{Equal contribution}\def\thefootnote{\arabic{footnote}}

\begin{abstract}
 Training of large-scale text-to-image and image-to-image models requires a huge amount of annotated data. While text-to-image datasets are abundant, data available for instruction-based image-to-image tasks like object addition and removal is limited. This is because of the several challenges associated with the data generation process, such as, significant human effort, limited automation, suboptimal end-to-end models, data diversity constraints and high expenses. We propose an automated data generation pipeline aimed at alleviating such limitations, and introduce GalaxyEdit - a large-scale image editing dataset for add and remove operations. We fine-tune the SD v1.5 model on our dataset and find that our model can successfully handle a broader range of objects and complex editing instructions, outperforming state-of-the-art methods in FID scores by 11.2\% and 26.1\% for add and remove tasks respectively. Furthermore, in light of on-device usage scenarios, we expand our research to include task-specific lightweight adapters leveraging the ControlNet-xs architecture. While ControlNet-xs excels in canny and depth guided generation, we propose to improve the communication between the control network and U-Net for more intricate add and remove tasks. We achieve this by enhancing ControlNet-xs with non-linear interaction layers based on Volterra filters. Our approach outperforms ControlNet-xs in both add/remove and canny-guided image generation tasks, highlighting the effectiveness of the proposed enhancement.
\end{abstract}
\vspace{-2ex}
\section{Introduction}
\begin{figure}
\centering
\includegraphics[width=0.9\linewidth]{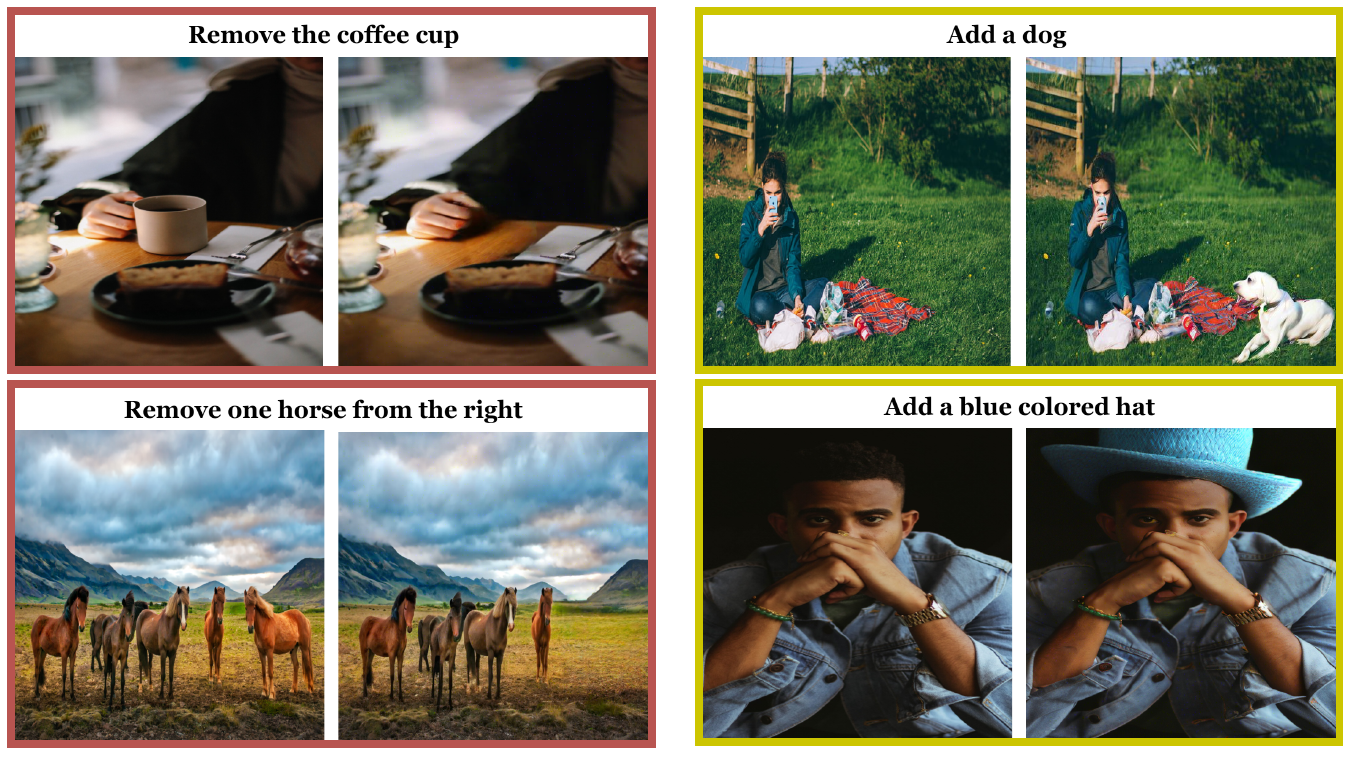}
\caption{Visual illustration of outputs generated by our proposed model.}
\vspace{-5mm}
\end{figure}

In the field of Generative AI, diffusion-based models have significantly advanced image editing tasks. Creating effective models requires high-quality, diverse datasets, but generating these at scale is challenging due to the labor-intensive nature of manual editing and annotation. This underscores the need for automated pipelines using Generative AI to efficiently produce datasets. Our focus is on object addition and removal tasks, which are crucial for various applications. Recent innovations have introduced mask-free methods \cite{ip2p, prompttoprompt, hive} alongside traditional mask-based techniques \cite{sdedit, blended_diffusion, imagen}. While mask-based methods are accurate, mask-free approaches offer the advantage of intuitive editing through natural language instructions. This allows users to specify changes without needing precise input of bounding boxes or masks, simplifying the process and making it more accessible to non-experts.

InstructPix2Pix (IP2P) \cite{ip2p}, a pioneering work in this domain, proposes a technique to generate an instruction based image editing dataset where source and target images are generated by a diffusion model using prompt-to-prompt \cite{prompttoprompt} technique. While a model trained on this dataset can handle a wide range of editing tasks, its ability to generalize to real-world images is restricted by the synthetic nature of the training image pairs. Another work Inst-Inpaint \cite{instinpaint} uses a mask based inpainting pipeline to generate data for instruction based object removal. A more recent work Paint-By-Inpaint Editing (PIPE) \cite{paintbyinpaint} generates data for object addition task by switching the source and target image pairs obtained by inpainting. Training on task-specific datasets for object addition and removal has proven highly effective, yielding significant improvements in model performance compared to general-purpose datasets. In this work we study model performance on object addition and removal tasks under varying levels of instruction complexity. Our approach is based on the insight that a single edit in the image domain can be articulated in multiple ways in the language domain. Building upon this observation, we have developed GalaxyEdit, an extensive dataset for object addition and removal tasks, where each pair of source and target images is accompanied by multiple instructions. This is achieved by considering a wide range of object categories, their attributes, and spatial locations to generate precise instructions. While most existing work in the literature deals with single object edits, in this work we also explore multi-object edits and present simple strategies to create a dataset for this purpose.

Training a successful image editing model requires maintaining a high level of fidelity to the original image while accurately executing the specified edit. Fine-tuning pre-trained models can adapt to specific tasks but risks catastrophic forgetting. A more efficient approach is to lock the parameters of the generative model and use a separate control network to guide the generation process \cite{t2i, controlnet, controlnetxs}. ControlNet-xs \cite{controlnetxs} achieves this by enabling bidirectional information flow between the encoder of the generative network and control network, resulting in a notable reduction in the size of the latter. Drawing inspiration from the success of ControlNet-xs in canny and depth-guided generation, we adapt it for instruction-based image editing. The utilization of natural images as a control signal adds to the complexity as the control network must provide both structural and semantic guidance to the generative network. By fixing the capacity of the control network, the quality of generation hinges on the efficiency of information exchange between the two networks. ControlNet-xs employs a straightforward information exchange scheme where intermediate features from one encoder are processed using a zero convolution block and then added to the output of the corresponding neural block in the other encoder. However, a limitation of this simple interaction scheme is that the control network is burdened with processing the input control signal while also facilitating information exchange in a manner that is comprehensible to both networks. To address this challenge, we propose delegating the task of information exchange to a small Volterra Neural Network (VNN), acting as a non-linear bridge between the two networks. VNN turns out to be the natural choice for non-linear information fusion as highlighted in \cite{vnnjmlr}. Moreover, VNNs have been successfully utilized in the past for tasks such as media restoration \cite{mrvnet}, facial recognition \cite{vnnface} and action recognition \cite{vnnaaai}. In our study, we evaluate the impact of adding a VNN based interaction layer for add/remove and canny to image generation tasks.

Here, we briefly summarise the key contributions made in this work. 1.) We present GalaxyEdit, a large-scale dataset designed for instruction-based image editing, enabling a wide variety of add and remove instructions. 2.) We propose ControlNet-Vxs, a new variant of ControlNet-xs utilizing Volterra Neural Networks to facilitate high capacity information exchange between the control network and pre-trained U-Net.
3.) Through extensive experimentation and ablation we demonstrate the efficacy of our dataset and superiority of our proposed model.

\section{Related Work}
Image editing can be broadly categorized into mask-based and mask-free approaches. Mask-based editing utilizes binary masks for structural guidance \cite{sdedit, blended_diffusion, imagen}. Conversely, mask-free editing, exemplified by Inpaint-Anything \cite{inpaintany} and DiffEdit \cite{diffedit}, offers flexibility through instructions, with recent advancements like HIVE \cite{hive} focusing on aligning instruction-based edits to user preferences.

\textbf{Image Control Guided Generation.} Recent works have focussed on utilizing image as a control signal to guide the pre-trained model generations. IP-Adapter \cite{ipadapter} utilizes a decoupled cross-attention strategy to guide the generation based on an image input. T2I-Adapter \cite{t2i} trains an additional network to modulate the intermediate feature maps of the pre-trained Stable Diffusion. ControlNet \cite{controlnet} replicates the encoder of pre-trained Stable Diffusion and links it to the base model using zero convolution initialized layers. ControlNet-xs \cite{controlnetxs} improves ControlNet by introducing bidirectional information flow between the encoders of the Control Network and Stable Diffusion.

\textbf{Volterra Neural Nets.} The Volterra Filter, originally introduced in \cite{volterra}, is used to approximate the non-linear relationship between input and the output. Unlike traditional activation functions, Volterra Filters incorporate non-linearities through higher-order convolutions, offering potential benefits in learning accurate non-linear functions in deep learning models. Previous works have introduced cascaded Volterra Networks that can be realistically implemented to perform highly non-linear tasks \cite{vnnjmlr, vnnaaai}.  A recent study, MR-VNET \cite{mrvnet}, incorporates VNN layers into the U-Net architecture to enhance performance in media restoration tasks.

\begin{figure*}[!ht]
    \includegraphics[scale=0.45]{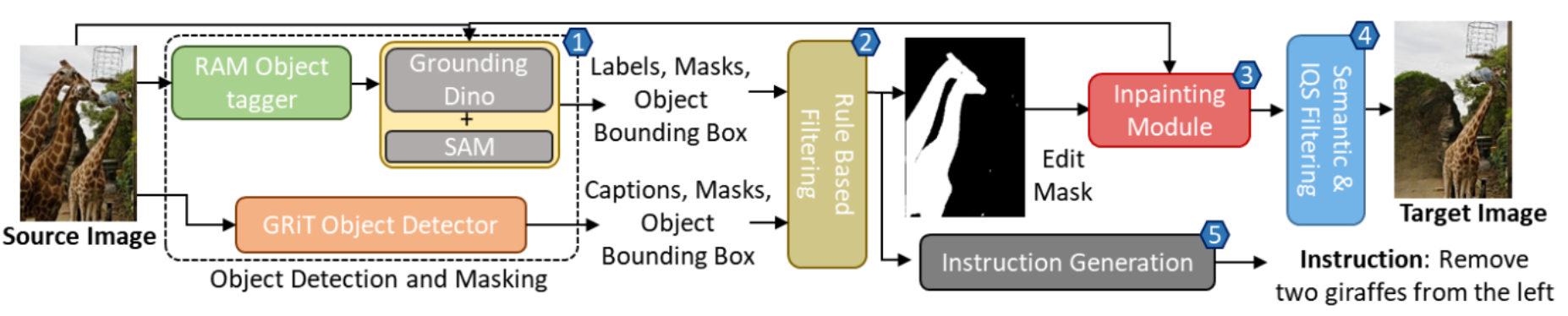}
  \caption{Proposed data generation pipeline for GalaxyEdit dataset.}
  \label{fig:datapipeline}
\end{figure*}

\textbf{Image Editing Datasets.} Dataset generation for image editing tasks has garnered a lot of attention in recent years. InstructPix2Pix \cite{ip2p} introduces a general purpose editing dataset comprising of synthetic source-target pairs along with edit instructions. Inst-Inpaint \cite{instinpaint} proposes the GQA-Inpaint dataset for object removal using a mask based inpainting. A more recent work PaintByInpaint \cite{paintbyinpaint} inverts the mask based inpainting process to generate image pairs for object addition and augments them with instructions generated using a VLM. Although these methods can generate high quality datasets, they focus only on single object edits and rely on external datasets to increase their diversity. For instance, PaintByInpaint utilizes the RefCOCO dataset \cite{refcoco} for referring expression based instrucions, InstInpaint relies on the scene graph annotations in GQA dataset. In this work, we provide a systematic approach to generate add-remove dataset with varying complexity of edits. The type of edits ranges from very simple/direct edits to more complex spatial and multi object edits. Spatial information has been used in Vision-Laguage datasets \cite{spatialvlm} to enhance their spatial reasoning capabilities, however such techniques haven't been utilized for image editing datasets.

\section{GalaxyEdit Dataset}
\label{dataset}
Our goal is to build a comprehensive dataset for add and remove tasks that supports edits on a wide range of objects using a diverse set of instructions. Towards this, we propose a unified pipeline comprising of 4 stages 

As a source image repository, we utilize the COCO dataset \cite{coco}, which contains 118k images annotated with object labels and masks. Fig \ref{fig:datapipeline} depicts the overall flow diagram for generating data using our pipeline. Given an input image, we select one or more objects for edit and obtain the inpainted image by running through the first three stages of our pipeline. The source and edit image pair obtained in this way serves as a training pair for the remove task. To obtain the training pair for the add task we simply swap the source and target images. Subsequently, we run the instruction generation module to generate a range of instructions for the edit, completing the data generation process. We refer the reader to the supplementary material for visualizations of the dataset samples and more details on the data generation process.  In the following sections we explain each module in detail.

\subsection{Object Detection and Mask Generation}

\textbf{Object Detection.} Our choice of object detector is guided by the diversity of objects we aim to accommodate and the nature of instructions we intend to create. To increase the range of detected objects we use an open set object tagger RAM \cite{ram}. The Recognize Anything Model (RAM) is an image tagging model that can tag any object category with high accuracy. The tags generated by RAM are then utilized by the GroundingDINO \cite{groundingdino} object detector. The combined detections from RAM and COCO annotations yield a total of $2887$ object categories. The detected bounding boxes serve as a source for mask generation and the associated labels serve as a useful resource for instruction generation. However, just simple labels are insufficient to create descriptive instructions. Hence, we adopt the GRiT \cite{grit} object detector for dense captioning of images. GRiT outputs attribute rich short captions for objects in images along with the detected bounding boxes and masks. We utilize these descriptive captions for instruction generation and masks for inpainting.

\textbf{Mask Generation.} The open set object detections from RAM require generation of masks for inpainting. We feed the bounding box outputs from GroundingDINO into the Segment Anything Model to obtain the masks for the detected objects. Since, the inpainting quality is largely determined by the quality of the masks we adopt some filtering criteria to filter out noisy and unethical detections which are discussed in the next section.

\subsection{Rule-Based and Semantic Filtering}
\textbf{Based on Object Size.} Objects that are either too small or too large are excluded from the inpainting process. Small objects are disregarded as they would have minimal impact on the inpainted result, while large objects are omitted to prevent unwanted artifacts from appearing in the final inpainting outcome. Specifically, we remove objects which are smaller than $0.18\%$ or larger than $50\%$ of the total image area.

\textbf{Based on Keywords.} Keyword-based filtering is implemented to eliminate unethical and incoherent object categories. Objects related to human body parts or clothing apparel, for instance, are excluded to address ethical considerations. Additionally, certain tags outputted by RAM may include verbs, generic nouns, or colors. While removing these could potentially impact the realism of the inpainting results, we choose to omit such detections to ensure the overall quality of the output.

\textbf{Semantic Filtering.} We utilize CLIP \cite{clip} to determine the similarity between the segmented region and its assigned object tag. Based on the similarity score we identify and exclude cases with masking errors or instances where the object is inadequately visible. To ensure accurate removal, we also validate that the similarity score of the segmented region post-removal is lower than the similarity score of the segmented region pre-removal. Further we employ classical image processing techniques to determine the quality of the inpainted region. Additional details are provided in the supplementary.

\subsection{Mask-Based Inpainting}
We use LaMa \cite{lama} as the inpainting model to remove the 
object of interest from the image. LaMa produces high quality inpainting results even for high image resolutions and large object masks. We also explore diffusion based inpainting method for removal. However, we find that the diffusion based method often hallucinates and leaves artifacts in the inpainted area. As shown in \cite{paintbyinpaint}, the use of such methods requires advanced methods of filtering. Before supplying the object masks to LaMa we first dilate the masks using a $15\times15$ kernel. This operation expands the mask at the object boundaries, thus compensating for any error that may have arisen during the mask generation process.

\subsection{Instruction Generation}
\label{subsec:inst-gen}
The final stage of our pipeline involves the generation of a diverse set of edit instructions that best align with the visual edit. We present four different instruction generation strategies each determined by the complexity of the instruction and editing operation.

\textbf{Simple Instructions.} These type of instructions follow a simple template such as \texttt{"add a <object>"} and \texttt{"remove the <object>"} for the corresponding add and remove tasks. The \texttt{<object>} tag is replaced with suitable object label obtained from RAM and COCO annotations. While it is convenient to use short instructions, the generic nature of such instructions might lead to ambiguity in edit tasks.

\textbf{Attribute-Based Instructions.} To add more specificity to the instructions we resort to addressing the object by it's attributes. Attributes refer to the color, size, shape, texture and other features of an object that help to distinguish one object from another. We prepare such kind of instructions by using attribute rich short captions from GRiT.
To these short captions we prepend the corresponding task prefix \texttt{"add"} or \texttt{"remove"}. While one can use an LLM to blend the task prefix with the short caption, we find that a simple rule based approach suffices to generate grammatically correct instructions.

\textbf{Spatial Instructions.} Often times it is desirable to refer to an object using it's relative spatial location, for e.g. \texttt{"<object1> to the left of <object2>"}, \texttt{"<object1> in front of <object2>"}. Generation of such instructions requires 3D spatial understanding. We achieve this by performing monocular depth estimation \cite{zoedepth} followed by a projection of 2D image into metric-scale 3D point cloud. This allows us to compare the relative positions of segmented objects in the pointcloud, based on their 3D bounding box coordinates. Given an object we find it's relative displacement with respect to an adjacent object in $x$, $y$ and $z$ directions. Based on this we assign predicates such as \texttt{left}, \texttt{right}, \texttt{above}, \texttt{below}, \texttt{front} and \texttt{behind}. To generate the instruction according to the template mentioned above, we uniquely refer \texttt{<object2>} using short attribute rich caption from GRiT and \texttt{<object1>} using the corresponding class label. To obtain the class label for the short captions we utilize Llama3, providing it with 3 in-context learning examples.

\textbf{Multi-Instance Instructions.} Editing multiple instances of an object in an image presents a significant challenge, exacerbated by the limited availability of relevant data for the task. We fill this gap by generating such data where more than one instance of an object are added or removed in an image. Towards this we first identify the instances of all objects of a class in an image. Next, we determine the layout of the instances based on the spread of the bounding boxes in $x$ and $y$ directions. Based on the inferred layout, we generate instruction of the form \texttt{"remove  k <objects> from <direction>"}. In the add case we drop the direction. The mask for each of these operations is formed by selecting the corresponding instance masks and combining them using the logical OR operation. The resulting mask is passed into the inpainting model to obtain the inpainted image for the generated instruction.

\section{Enhanced Diffusion Adapter}

\subsection{Problem Formulation}
Given the generated GalaxyEdit dataset with input images $\boldsymbol{X_I}=x_I^1, x_I^2, ..., x_I^N$, edit instructions $\boldsymbol{E}=e^1,e^2, ...e^N$, the objective is to learn an editing function $\mathcal{G}$ to generate a set of edited images, $\boldsymbol{X_E} = x_E^1, x_E^2, ..., x_E^N$. Specifically, we are interested in a class of functions where the output of a frozen base model $\mathcal{F}_b(x_b; \Theta_b)$ is modulated using a control network $\mathcal{F}_c(x_c; \Theta_c)$ through an interaction function $\Phi(., .)$. The generalised form of such functions can be expressed as follows
\begin{equation}
\mathcal{G} = \Phi(\mathcal{F}_b, \mathcal{F}_c)
\end{equation}

Let $\mathcal{F}_b^{(i)}(x_b; \Theta_b)$ be the $i^{th}$ neural block in the pre-trained base model that processes an input feature map $x_b^{(i-1)}$ to produce $x_b^{(i)}$.
\begin{equation}
x_b^{(i)} = \mathcal{F}_b^{(i)}(x_b^{(i-1)}; \Theta_b)
\end{equation}
ControlNet \cite{controlnet} influences the output of the pre-trained neural block by connecting a trainable copy of the corresponding block through zero convolution layer. Subsequently, the interaction function in ControlNet computes the block output as shown in equation \ref{eqn:infusion2base}.
\begin{equation}
\label{eqn:infusion2base}
\begin{split}
x_b^{(i)} = \Phi(\mathcal{F}_b^{(i)}, \mathcal{F}_c^{(i)}) & =  \mathcal{F}_b^{(i)}(x_b^{(i-1)}; \Theta_b) \\ & + \mathcal{Z}_{bc}^{(i)}(\mathcal{F}_c^{(i)}(x_c^{(i-1)}; \Theta_c);\Theta_z)
\end{split}
\end{equation}
where $\mathcal{Z}_{bc}^{(i)}(.;.)$ is a zero initialized first order convolution connecting control to base. 

Whereas, ControlNet-xs \cite{controlnetxs} features a bidirectional information exchange system in which the control network influences the base network (Eqn. \ref{eqn:infusion2base}), while also receiving information from the base network as follows,

\begin{equation}
\label{eqn:infusion2control}
\begin{split}
x_c^{(i)} = \Phi(\mathcal{F}_c^{(i)}, \mathcal{F}_b^{(i)}) & =  \mathcal{F}_c^{(i)}(x_c^{(i-1)}; \Theta_c) \\ & + \mathcal{Z}_{cb}^{(i)}(\mathcal{F}_b^{(i)}(x_b^{(i-1)}; \Theta_b);\Theta_z)
\end{split}
\end{equation}

However, both methods are limited by the linear feature fusion used in the interaction function $\Phi(.,.)$. In contrast, we propose the use of second order volterra filter as a non-linear information bridge between the base and control network.

\begin{figure}
\centering
\includegraphics[scale=0.5]{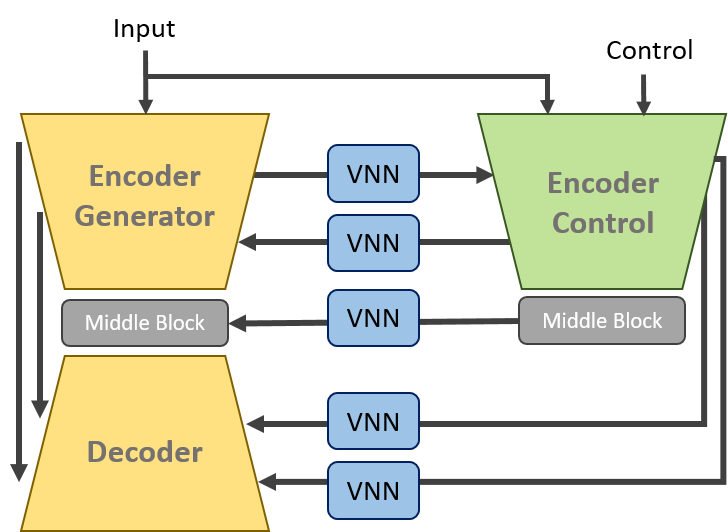}
\caption{\textbf{ControlNet-Vxs Model Architecture.}}
\label{fig:cnetvxs}
\vspace{-5mm}
\end{figure}

\subsection{Volterra Filters: A solution to optimal Information Exchange}
In our proposed methodology, we utilize a Volterra Neural Network (VNN) to implement the interaction function. A Volterra layer is defined based on the second-order approximation of the Volterra Series, with the second-order kernel implemented using the lossy approximation method from \cite{mrvnet}. The output of the $l^{th}$ Volterra layer in a VNN is then calculated as follows,

\begin{equation}
\begin{split}
V^{(l)}(X^{(l-1)}) & = \boldsymbol{W^1} \circledast X^{(l-1)} \\ & + \sum_{q=1}^Q\boldsymbol{W^2}_{aq} \circledast X^{(l-1)}\cdot\boldsymbol{W^2}_{bq} \circledast X^{(l-1)}
\end{split}
\end{equation}

where, $Q$ represents the rank of the approximation.

We incorporate the enhanced interaction function in ControlNet-xs and propose a new variant ControlNet-Vxs. In our approach, the feature map from the control block is fused into the feature map of the base neural block through a VNN layer.  The non linear fusion output from the VNN is merged with the base model features by taking a weighted combination of the two. Mathematically, the fusion into the base neural block can be expressed as

\begin{equation}
\label{eqn:infusion2basevnn}
\begin{split}
x_b^{(i)} = \Phi(\mathcal{F}_b^{(i)}, \mathcal{F}_c^{(i)}) & =  (1-w^{(i)}) \cdot \mathcal{F}_b^{(i)}(x_b^{(i-1)}; \Theta_b) \\ & + w^{(i)} \cdot V_{bc}^{(i)}([\mathcal{F}_{b}^{(i)}, \mathcal{F}_{c}^{(i)}];\Theta_z)
\end{split}
\end{equation}

where $[.,.]$ represents the concatenation of features along the channel dimension, $w^{(i)}$ is the learnable fusion weight, $V_{bc}(.;.)$ is a zero initialized VNN layer connecting control to the base. 

Similarly, the fusion from the base neural block into the control block can be defined as follows:
\begin{equation}
\label{eqn:infusion2controlvnn}
\begin{split}
x_c^{(i)} = \Phi(\mathcal{F}_c^{(i)}, \mathcal{F}_b^{(i)}) & =  (1-w^{(i)}) \cdot \mathcal{F}_c^{(i)}(x_c^{(i-1)}; \Theta_c) \\ & + w^{(i)} \cdot V_{cb}^{(i)}([\mathcal{F}_{c}^{(i)}, \mathcal{F}_{b}^{(i)}];\Theta_z)
\end{split}
\end{equation}

While we represent equations \ref{eqn:infusion2basevnn}, \ref{eqn:infusion2controlvnn} using a single layer of VNN for the ease of notation, multiple layers can be cascaded to realize more complex non linear fusion functions.

\begin{figure*}[!ht]

\begingroup

\setlength{\tabcolsep}{1.5pt}
\centering
\begin{tabular}{cccccccc}
\multicolumn{1}{c}{\centering\textbf{Source}} & \multicolumn{1}{c}{\centering\textbf{IP2P}} & \multicolumn{1}{c}{\centering\textbf{Inst-Inpaint}} & \multicolumn{1}{c}{\centering\textbf{GalaxyEdit}} & \multicolumn{1}{c}{\centering\textbf{Source}} & \multicolumn{1}{c}{\centering\textbf{IP2P}} & \multicolumn{1}{c}{\centering\textbf{PIPE}} & \multicolumn{1}{c}{\centering\textbf{GalaxyEdit}}\\ 
\includegraphics[height=0.12\textwidth, width=0.12\textwidth]{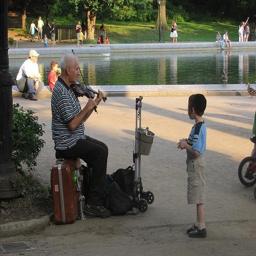} &
\includegraphics[height=0.12\textwidth, width=0.12\textwidth]{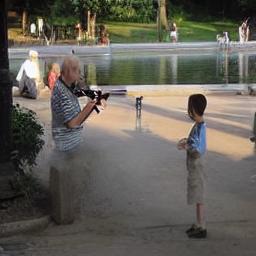}   &
\includegraphics[height=0.12\textwidth, width=0.12\textwidth]{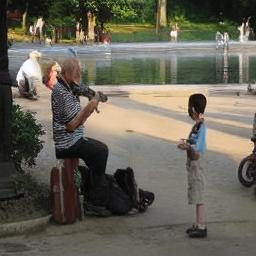}   &
\includegraphics[height=0.12\textwidth, width=0.12\textwidth]{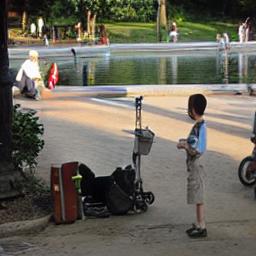} & 
\includegraphics[height=0.12\textwidth, width=0.12\textwidth]{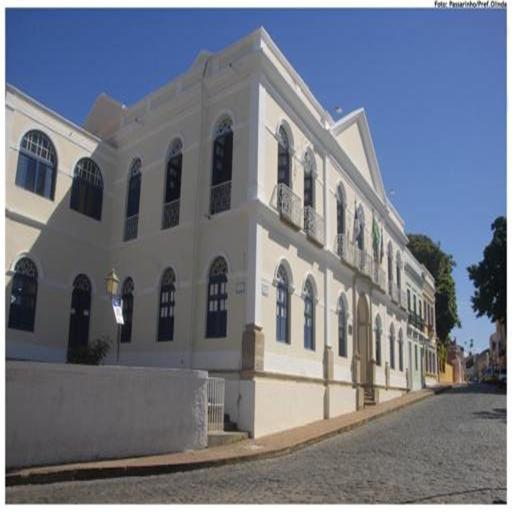} &
\includegraphics[height=0.12\textwidth, width=0.12\textwidth]{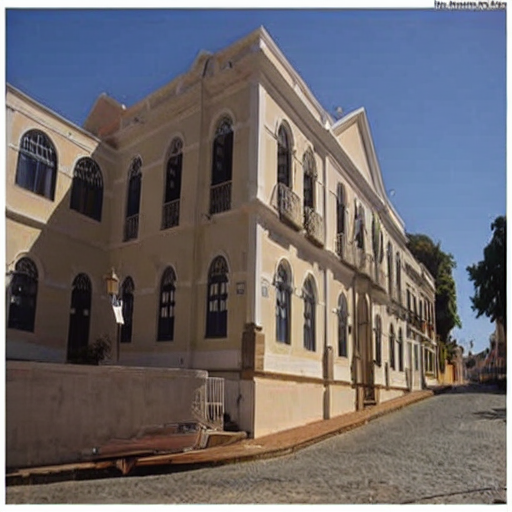}   &
\includegraphics[height=0.12\textwidth, width=0.12\textwidth]{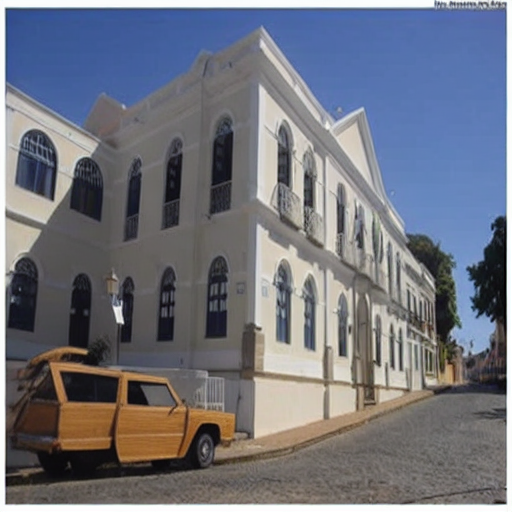}   &
\includegraphics[height=0.12\textwidth, width=0.12\textwidth]{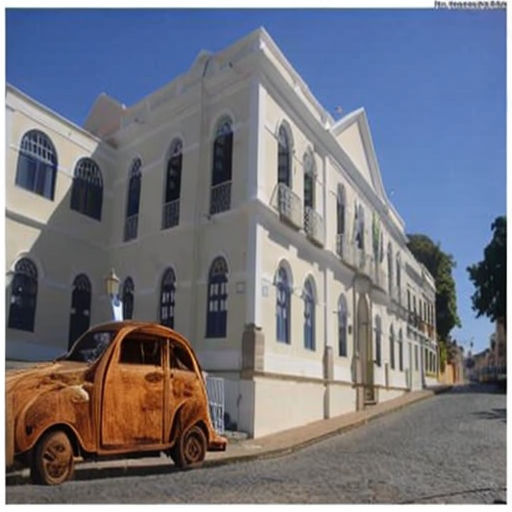}\\
\multicolumn{4}{c}{\centering Remove the person playing a violin} & \multicolumn{4}{c}{\centering Add a wooden vintage car}\\
\includegraphics[height=0.12\textwidth, width=0.12\textwidth]{imgs/results/sd15/remove_the_dark_brown_cow/src.jpg} &
\includegraphics[height=0.12\textwidth, width=0.12\textwidth]{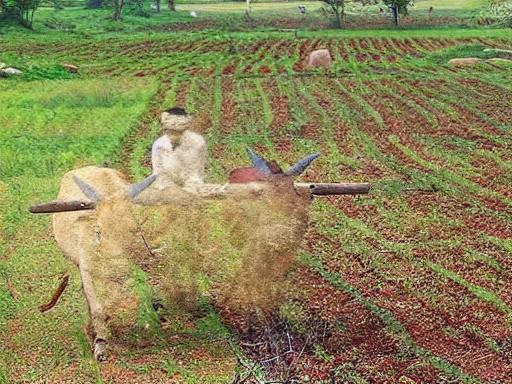}  &
\includegraphics[height=0.12\textwidth, width=0.12\textwidth]{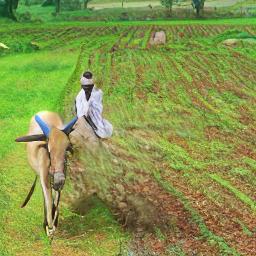}   &
\includegraphics[height=0.12\textwidth, width=0.12\textwidth]{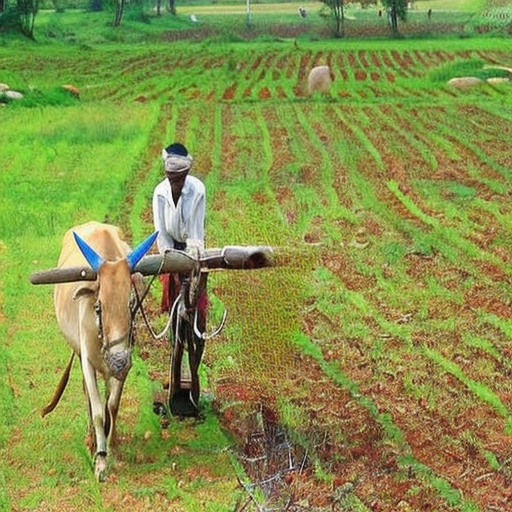} &
\includegraphics[height=0.12\textwidth, width=0.12\textwidth]{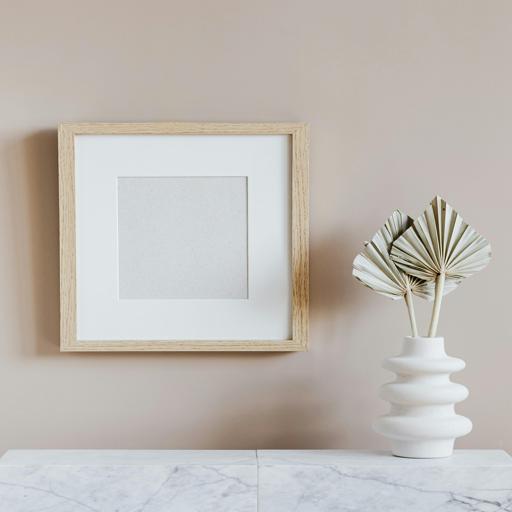} &
\includegraphics[height=0.12\textwidth, width=0.12\textwidth]{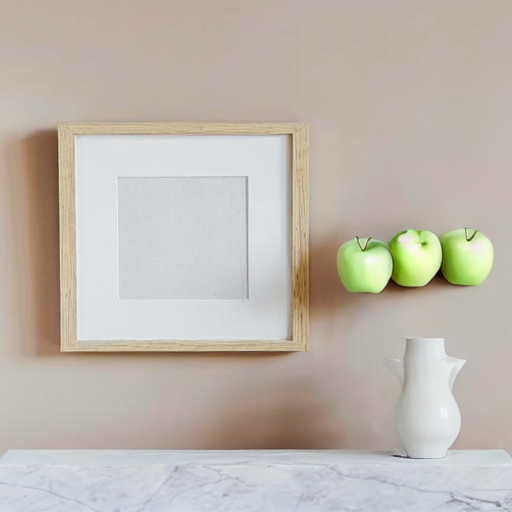}   &
\includegraphics[height=0.12\textwidth, width=0.12\textwidth]{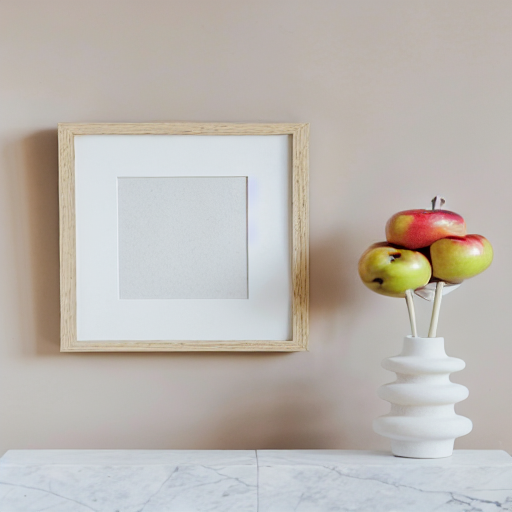}   &
\includegraphics[height=0.12\textwidth, width=0.12\textwidth]{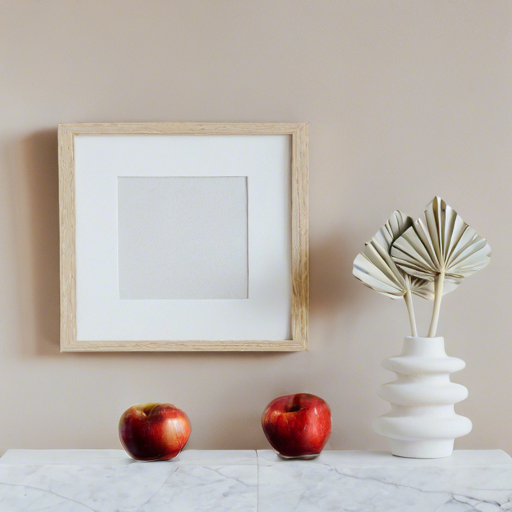}\\
\multicolumn{4}{c}{\centering Remove the dark brown cow} & \multicolumn{4}{c}{\centering Add two apples}\\
\includegraphics[height=0.12\textwidth, width=0.12\textwidth]{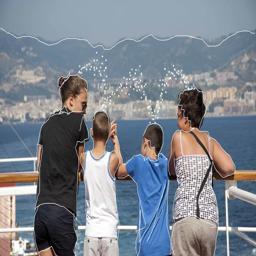}  &
\includegraphics[height=0.12\textwidth, width=0.12\textwidth]{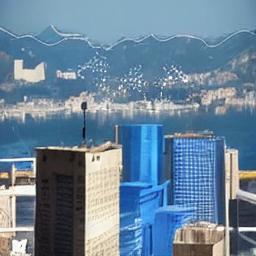} &
\includegraphics[height=0.12\textwidth, width=0.12\textwidth]{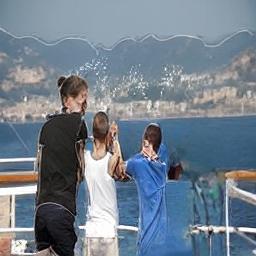} &
\includegraphics[height=0.12\textwidth, width=0.12\textwidth]{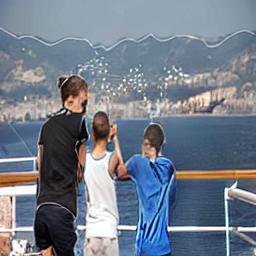} & 
\includegraphics[width=0.12\textwidth]{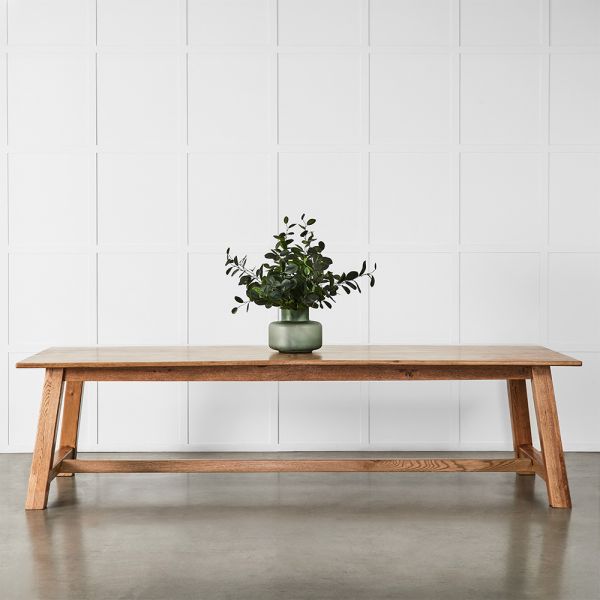} &
\includegraphics[height=0.12\textwidth, width=0.12\textwidth]{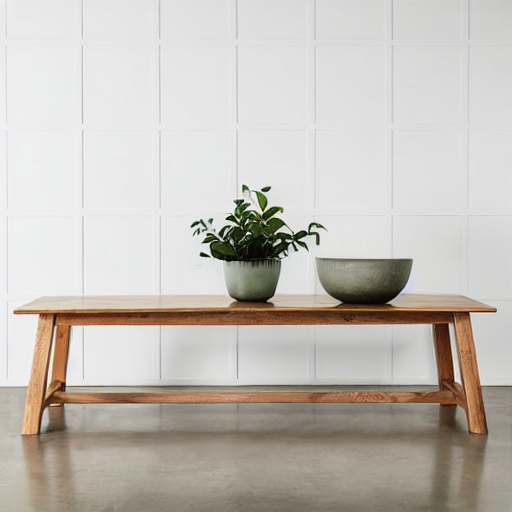}   &
\includegraphics[height=0.12\textwidth, width=0.12\textwidth]{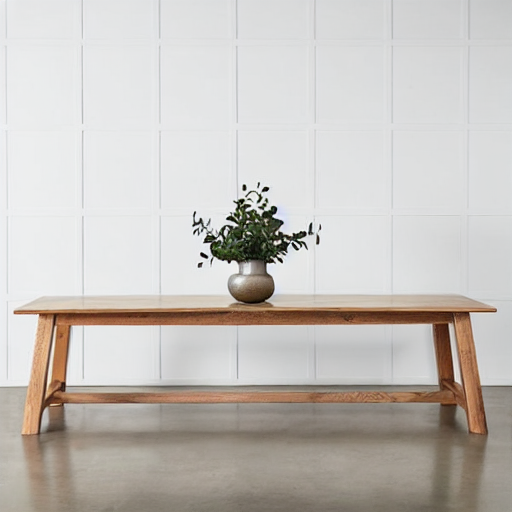}   &
\includegraphics[height=0.12\textwidth, width=0.12\textwidth]{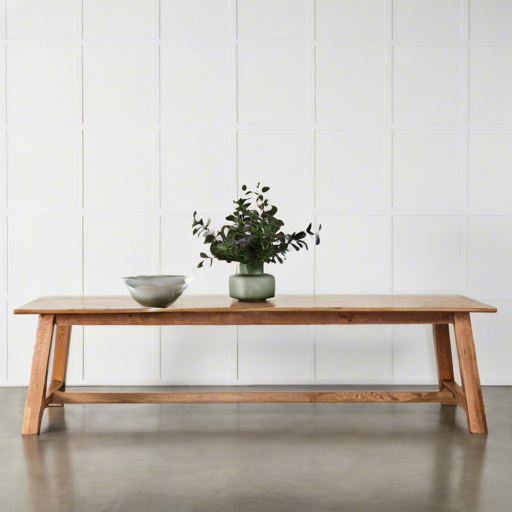} \\
\multicolumn{4}{c}{\centering Remove the person to the right of the person in blue shirt} & \multicolumn{4}{c}{\centering Add a bowl to the left of the potted plant}\\
\includegraphics[height=0.12\textwidth, width=0.12\textwidth]{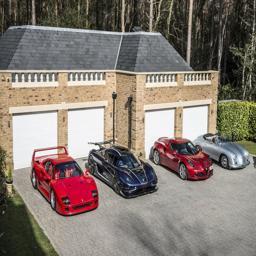}  &
\includegraphics[height=0.12\textwidth, width=0.12\textwidth]{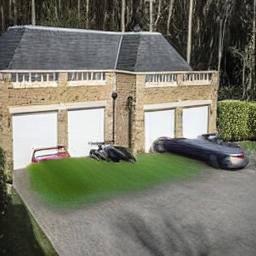} &
\includegraphics[height=0.12\textwidth, width=0.12\textwidth]{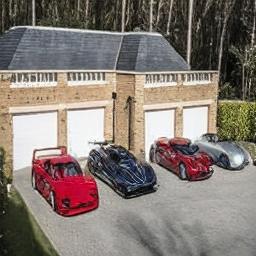}   &
\includegraphics[height=0.12\textwidth, width=0.12\textwidth]{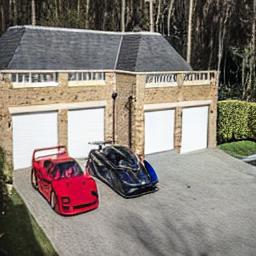} &
\includegraphics[width=0.12\textwidth]{imgs/results/sd15/grid10/src_207.jpg} &
\includegraphics[height=0.12\textwidth, width=0.12\textwidth]{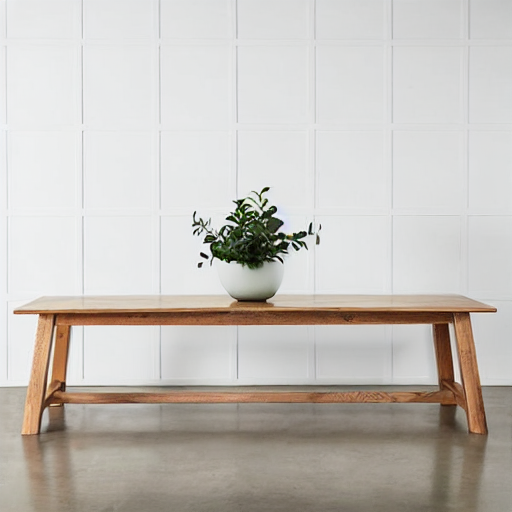}   &
\includegraphics[height=0.12\textwidth, width=0.12\textwidth]{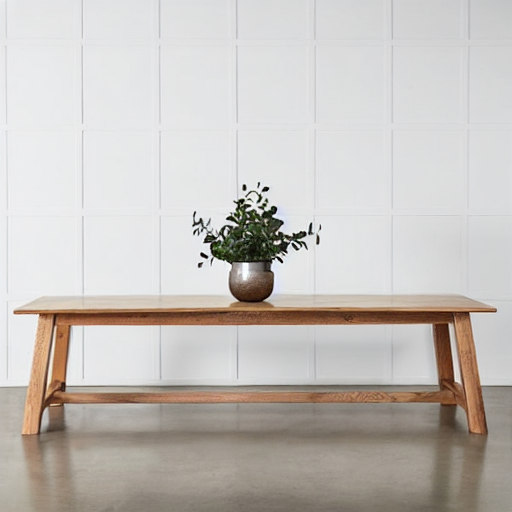}   &
\includegraphics[height=0.12\textwidth, width=0.12\textwidth]{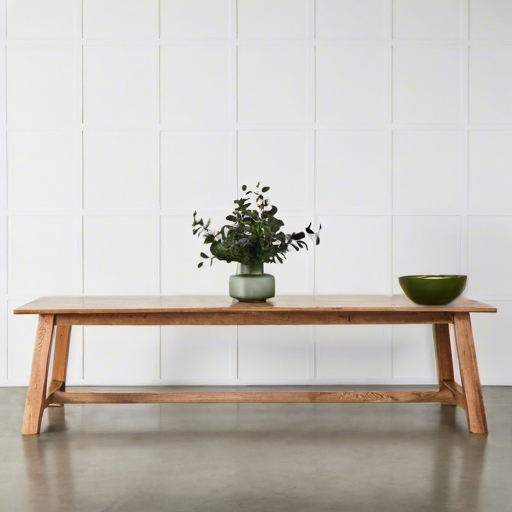} \\
\multicolumn{4}{c}{\centering Remove two cars from the right} & \multicolumn{4}{c}{\centering Add a bowl to the right of the potted plant} \\
\includegraphics[height=0.12\textwidth, width=0.12\textwidth]{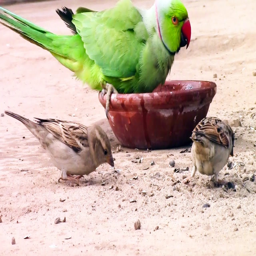}  &
\includegraphics[height=0.12\textwidth, width=0.12\textwidth]{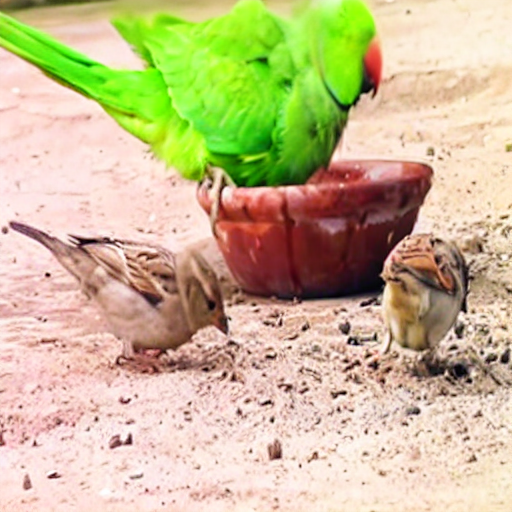}   &
\includegraphics[height=0.12\textwidth, width=0.12\textwidth]{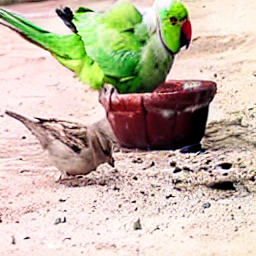}  &
\includegraphics[height=0.12\textwidth, width=0.12\textwidth]{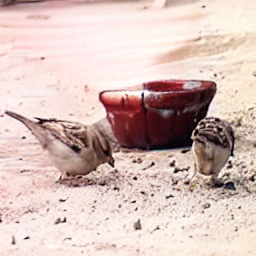} &
\includegraphics[height=0.12\textwidth, width=0.12\textwidth]{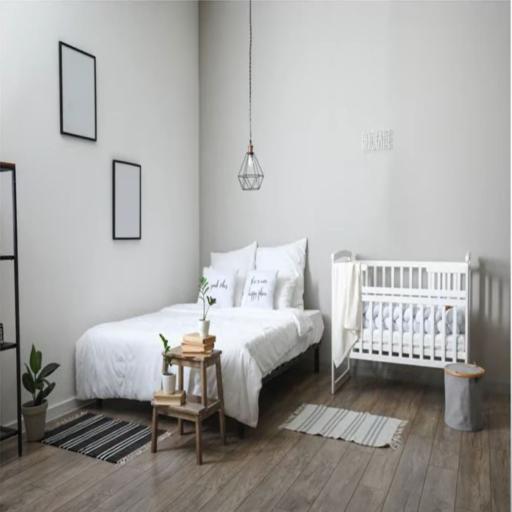} &
\includegraphics[height=0.12\textwidth, width=0.12\textwidth]{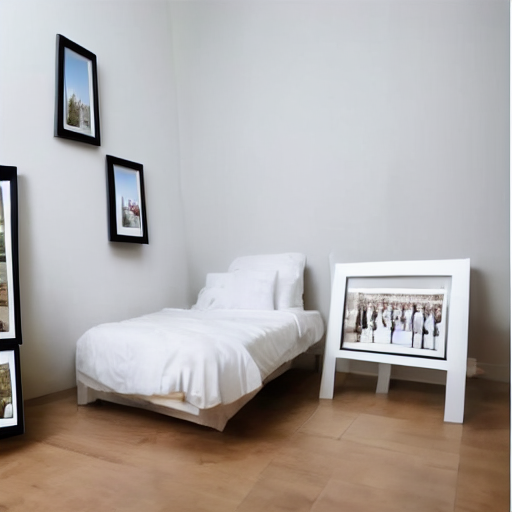}   &
\includegraphics[height=0.12\textwidth, width=0.12\textwidth]{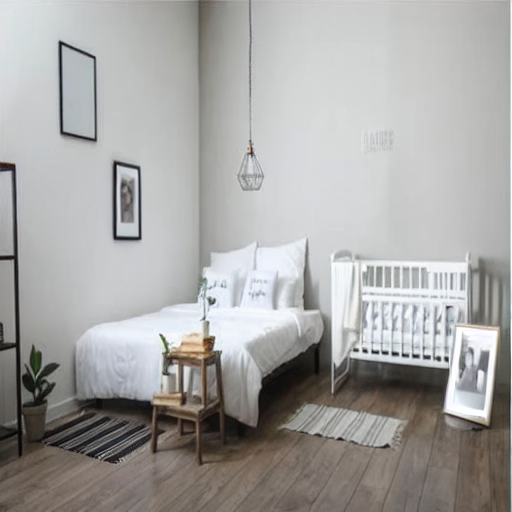}   &
\includegraphics[height=0.12\textwidth, width=0.12\textwidth]{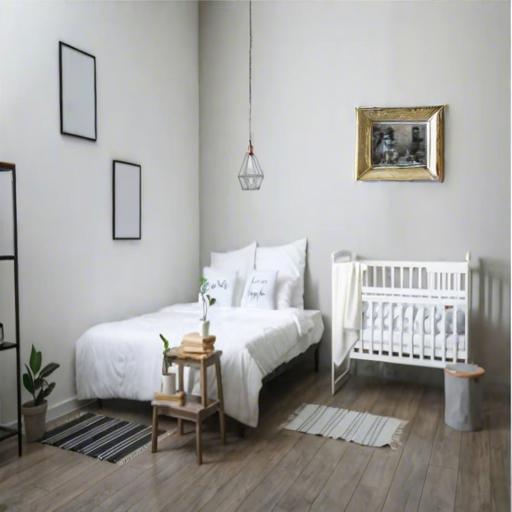} \\
\multicolumn{4}{c}{\centering Remove the parrot} & \multicolumn{4}{c}{\centering Add a photo frame} \\
\end{tabular}
\endgroup
\caption{\textbf{Qualitative results on Object Removal (Left) and Object Addition (Right).} We compare our model against leading methods for add and remove showcasing diverse instruction following capability while maintaining high fidelity to the input image.}
\label{fig:qual-res-sd}
\end{figure*}

\begin{figure}[ht]
\centering
\subfloat[Object Removal]{\includegraphics[width=0.49\linewidth]{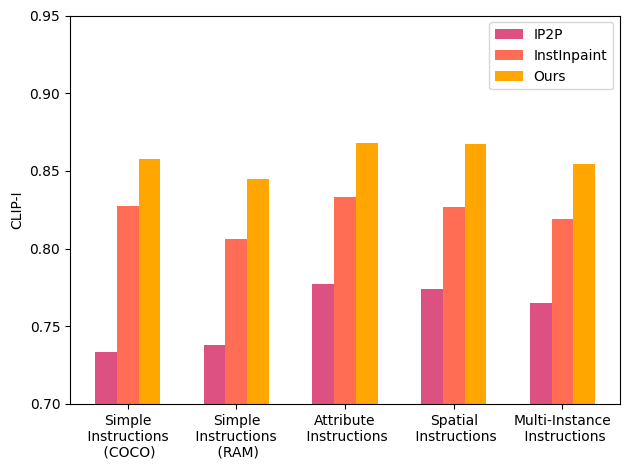}}
\hfill
\subfloat[Object Addition]{\includegraphics[width=0.49\linewidth]{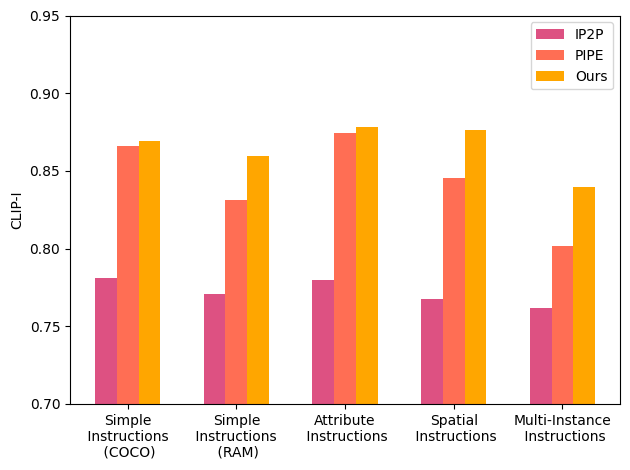}}
\caption{Category-wise results on GalaxyEdit test set}
\label{fig:cat-wise-gedit}
\end{figure}

\subsection{Model Training}
We adopt the same training recipe as detailed in \cite{controlnetxs}. The control model parameters are initialized randomly while the base model parameters are frozen. In all our experiments we use Stable Diffusion v1.5 as our base model. We train our models on two different conditionings - canny conditioning for canny to image generation task and image conditioning for the add/remove tasks. Our model is trained using the standard diffusion objective:

\begin{equation}
\mathcal{L} = \mathbb{E}_{z_0, t,c_c, c_t, \epsilon \sim \mathcal{N}(0, 1)}\norm{\epsilon - \epsilon_{\theta}(z_t, t, c_c, c_t)}_2^2
\end{equation}

where, $z_0$ represents the target image latent, $z_t$ represents the noisy latent at time step $t$, $c_c$ represents the control conditioning and $c_t$ represents the text conditioning.

\section{Experiments}

\subsection{Dataset Quality Assessment}
\label{sec:dqa}
\textbf{Experimental Setup.} We fine-tune an SD v1.5 model using $800K$ samples from the GalaxyEdit dataset and compare its performance against models trained on benchmark add/remove datasets. In our comparisons, we consider PIPE \cite{paintbyinpaint} and InstInpaint \cite{instinpaint} as baselines for object addition and removal tasks, and IP2P \cite{ip2p} as a general-purpose editing model for both scenarios. InstInpaint trains a low capacity LDM model whereas, PIPE and IP2P fine-tune the SDv.15 model. To ensure consistency across methods, we fine-tune an SD v1.5 model on each of these datasets following the training procedure described in \cite{ip2p}.

\begin{table*}[!ht]
\centering
\resizebox{0.85\textwidth}{!}{
\begin{tabular}{lcccccc|cccccc}
\hline
\multirow{2}{*}{Methods} & \multicolumn{6}{c|}{REMOVE}                                                                                & \multicolumn{6}{c}{ADD}                                                                                    \\ \cline{2-13} 
                         & L1 $\downarrow$             & L2 $\downarrow$             & CLIP-T         $\downarrow$ & CLIP-I   $\uparrow$       & DINO $\uparrow$            & FID  $\downarrow$            & L1 $\downarrow$             & L2       $\downarrow$       & CLIP-T  $\uparrow$         & CLIP-I $\uparrow$          & DINO $\uparrow$            & FID     $\downarrow$         \\ \hline
IP2P                     & 0.1331          & 0.0436          & 0.1878          & 0.7573          & 0.6959          & 59.880          & 0.1348          & 0.0471          & \textbf{0.2102} & 0.7719          & 0.734          & 72.390          \\
Inst-Inpaint             & 0.0757          & 0.0183          & 0.1862          & 0.8222          & 0.8455         & 50.584          & -               & -               & -               & -               & -               & -                \\
PIPE                     & -               & -               & -               & -               & -               & -                & 0.0749          & 0.0247          & 0.2032          & 0.8427          & .8872          & 54.633         \\
GalaxyEdit                     & \textbf{0.0605} & \textbf{0.0138} & \textbf{0.1821} & \textbf{0.8583} & \textbf{0.8948} & \textbf{37.343} & \textbf{0.0650} & \textbf{0.0182} & 0.2017          & \textbf{0.8647} & \textbf{0.9188} & \textbf{48.508} \\ \hline
\end{tabular}
}
\caption{\textbf{Results on GalaxyEdit test set.} (-) indicates method not applicable to the task.}
\label{tab:dqa-gedit}
\end{table*}

\begin{table*}[!ht]
\centering
\resizebox{0.85\textwidth}{!}{
\begin{tabular}{lcccccc|cccccc}
\hline
\multirow{2}{*}{Methods} & \multicolumn{6}{c|}{REMOVE}                                                                                & \multicolumn{6}{c}{ADD}                                                                                    \\ \cline{2-13} 
                          & L1 $\downarrow$             & L2 $\downarrow$             & CLIP-T         $\downarrow$ & CLIP-I   $\uparrow$       & DINO $\uparrow$            & FID  $\downarrow$            & L1 $\downarrow$             & L2       $\downarrow$       & CLIP-T  $\uparrow$         & CLIP-I $\uparrow$          & DINO $\uparrow$            & FID     $\downarrow$             \\ \hline
IP2P                     & 0.1227          & 0.04            & 0.1833          & 0.7456          & 0.496          & 167.9966         & 0.106           & 0.0339          & \textbf{0.2159} & 0.8012          & 0.5903          & 114.677         \\
Inst-Inpaint             & 0.0963          & 0.0329          & \textbf{0.1776} & 0.809           & 0.5993          & 104.434         & -               & -               & -               & -               & -               & -                \\
PIPE                     & -               & -               & -               & -               & -               & -                & 0.0783          & 0.0273           & 0.2038          & 0.8374          & 0.6988         & 96.546         \\
GalaxyEdit                     & \textbf{0.0804} & \textbf{0.0267} & 0.1787          & \textbf{0.8408} & \textbf{0.6712} & \textbf{96.904} & \textbf{0.0686} & \textbf{0.0212} & 0.1967          & \textbf{0.8535} & \textbf{0.7245} & \textbf{85.184} \\ \hline
\end{tabular}
}

\caption{\textbf{Results on MagicBrush test set.} (-) indicates method not applicable to the task.}
\label{tab:dqa-mb}
\end{table*}

\begin{table*}[!ht]
\centering
\resizebox{0.85\textwidth}{!}{
\begin{tabular}{lcccccc|cccccc}
\hline
\multirow{2}{*}{Methods} & \multicolumn{6}{c|}{REMOVE}                                                                                & \multicolumn{6}{c}{ADD}                                                                                    \\ \cline{2-13} 
                          & L1 $\downarrow$             & L2 $\downarrow$             & CLIP-T         $\downarrow$ & CLIP-I   $\uparrow$       & DINO $\uparrow$            & FID  $\downarrow$            & L1 $\downarrow$             & L2       $\downarrow$       & CLIP-T  $\uparrow$         & CLIP-I $\uparrow$          & DINO $\uparrow$            & FID     $\downarrow$             \\ \hline
ControlNet-xs                     & 0.071          & 0.0176            & 0.1816          & 0.8378          & 0.8676          & 43.691         & 0.0824           & 0.0227          & 0.199 & 0.8315          & 0.8882          & 57.902         \\
ControlNet-Vxs                     & \textbf{0.0638} & \textbf{0.0144} & \textbf{0.18}          & \textbf{0.8491} & \textbf{0.89} & \textbf{38.686} & \textbf{0.0808} & \textbf{0.0227} & \textbf{0.2005} & \textbf{0.8442} & \textbf{0.8995} & \textbf{55.798} \\ \hline
\end{tabular}
}

\caption{\textbf{Performance of Adapter Based Methods on GalaxyEdit test set.} Comparison of our model and ControlNet-xs trained on the GalaxyEdit dataset.}
\label{tab:cnet-gedit}
\vspace{-3mm}
\end{table*}

\textbf{Datasets.} Evaluation of the fine-tuned models is carried out on the following test sets - (i) GalaxyEdit Test Set: A test set consisting of $1000$ source-target-instruction pairs generated using the pipeline discussed in Section \ref{dataset}, from $2000$ held-out images from the COCO validation split. (ii) MagicBrush Test Set \cite{magicbrush}: We filter $130$ edits for the add task and $62$ edits for the remove task. Additional details have been provided in the supplementary.

\textbf{Metrics.} In our quality estimation process, we leverage a combination of model-free and model-based metrics. Specifically, for the model-free metrics, we calculate the $L_1$ and $L_2$ distances between the edited and reference images. For the model-based metrics, we integrate CLIP-I based on CLIP embeddings \cite{clip} and DINO \cite{dino} to evaluate the semantic similarity between the edited and reference images. We also use CLIP-T \cite{clipscore} to assess the alignment between the edited image and textual instructions, and FID \cite{fid} to measure the distance between the distributions of reference and edited images.

\subsection{Adapter Evaluation}

In our experiments, we employ the Type-B variant of ControlNet-xs \cite{controlnetxs} containing 55 million parameters. 

We enhance this variant to create ControlNet-Vxs by replacing the linear information transfer in ControlNet-xs with VNN layers that connect encoders of the base and control models. We evaluate and compare the performance of our proposed model, ControlNet-Vxs, against ControlNet-xs on two specific tasks.

\textbf{Instruction Based Image Editing.} We train both models on the GalaxyEdit dataset and evaluate on the GalaxyEdit test set. During training, the original image is given as a control input to the model while the edited image serves as the target. Our evaluation includes the same metrics outlined in section \ref{sec:dqa}.

\textbf{Canny to Image Generation.} For this task we use 1 million images from the LAION \cite{laion} dataset. During training, we extract canny edges from an image using random thresholds, which are then supplied to the model as control inputs. For evaluation, we consider several metrics including CLIP-T for image-text similarity \cite{clipscore}, CLIP aesthetic scores CLIP-Aes \cite{clipaes}, perceptual similarity scores using LPIPS \cite{lpips}, and FID for distribution distance measurement \cite{fid}.
\section{Results}

In this section, we attempt to answer the following research questions (\textbf{RQs}).
\begin{itemize}
\item[] \textbf{RQ1.} How does the performance of a diffusion model trained on the GalaxyEdit dataset compare to that of the same model trained on other add/remove datasets? (\ref{res:rq1})
\item[] \textbf{RQ2.} Does a model trained on our dataset generalize to publicly available datasets? (\ref{res:rq2})
\item[] \textbf{RQ3.} Does ControlNet-Vxs improve quality of generated images? (\ref{res:rq3})
\end{itemize}

\subsection{Performance Comparison on GalaxyEdit}
\label{res:rq1}
In this section, we evaluate the quality of the proposed GalaxyEdit dataset. For this purpose, we fine-tune the Stable Diffusion v1.5 on benchmark add/remove datasets and the GalaxyEdit dataset. The results are presented in Table \ref{tab:dqa-gedit}. We find that our model significantly outperforms the baselines in both add and remove tasks. IP2P being a general purpose editing model underperforms in both the tasks. One obvious reason for the improvement seen in the remove task is the sheer scale of our dataset. Extensive object tagging, inpainting and diverse instruction generation capabilities of our proposed pipeline, allow us to generate a large scale dataset from a very limited set of images. Significant gains are seen on the add task as well. However, to understand the areas where our model outperforms the baselines, we conduct split-wise evaluations (Fig \ref{fig:cat-wise-gedit}). Across all categories, our model consistently outperforms the baselines in the remove task, demonstrating the benefits of training on a large-scale dataset with a diverse range of instructions. In the object addition task, we observe that the performance of the PIPE model closely matches with our model in simple and attribute-rich instructions, but falls short in others. Notably, all three models exhibit subpar performance when handling multi-instance instructions compared to other types of instructions. Our overall observation indicates that performance tends to be relatively better for instruction categories that are more specific, such as spatial and attribute-based instructions.

\textbf{Human Evaluation.} To assess the quality and efficacy of our proposed dataset, we conducted a human evaluation of both the dataset and the outputs generated by different models. We selected 100 random samples from the GalaxyEdit test set and generated outputs using various models. A panel of 25 human evaluators was then requested to evaluate these outputs. Each evaluator rated the model outputs on a scale of 1 to 5, considering the input image and edit instructions, without knowing which model produced each output. Additionally, evaluators rated the quality of the ground truth images. Overall the evaluators noted a better instruction following capability in our model as evident from the results reported in Table \ref{tab:human_eval}.

\begin{table}[htb]
\centering
\resizebox{0.91\linewidth}{!}{
\begin{tabular}{lcc}
\hline
                            & \multicolumn{2}{l}{Average Rating}                   \\ \hline
                            & \multicolumn{1}{l}{Remove} & \multicolumn{1}{l}{Add} \\ \hline
IP2P                        & 1.6                      & 1.9                   \\
Inst-Inpaint                & 2.9                      & -                       \\
PIPE                        & -                          & 3.8                   \\
GalaxyEdit                  & \textbf{4.0}             & \textbf{4.1}          \\ \hline
Average Ground Truth Rating & 4.8                      & \multicolumn{1}{l}{}    \\ \hline
\end{tabular}
}
\caption{\textbf{Human Evaluation.} Average ratings out of 5, rounded to the first decimal.}
\label{tab:human_eval}
\vspace{-5mm}
\end{table}

\subsection{Generalization to External Datasets}
\label{res:rq2}
We further assess the generalization capacity of a model trained on our dataset to publicly available datasets. For this we train all models on their respective datasets and evaluate on the MagicBrush test set. As shown in Table \ref{tab:dqa-mb}, our model outperforms the baseline methods on 4 out of 5 metrics. Owing to the diverse nature of training instructions, our model shows better generalization capabilities compared to the baselines. Moreover, the qualitative results illustrated in Figure \ref{fig:qual-res-sd} demonstrate that our model effectively learns to execute user-provided instructions of varying complexities with high accuracy.

\begin{figure}[!ht]
\centering
\begingroup
\setlength{\tabcolsep}{1.5pt}
\resizebox{0.8\linewidth}{!}{
\begin{tabular}{ccc}
\multicolumn{1}{c}{\centering\textbf{Source}} & \multicolumn{1}{c}{\centering\textbf{ControlNet-xs}} & \multicolumn{1}{c}{\centering\textbf{ControlNet-Vxs}}\\ 
\includegraphics[height=0.3\linewidth, width=0.3\linewidth]{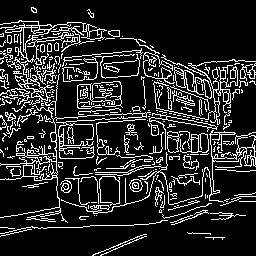} &
\includegraphics[height=0.3\linewidth, width=0.3\linewidth]{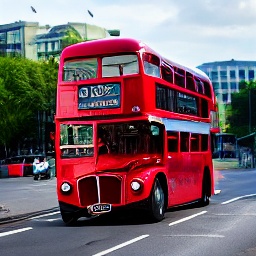}   &
\includegraphics[height=0.3\linewidth, width=0.3\linewidth]{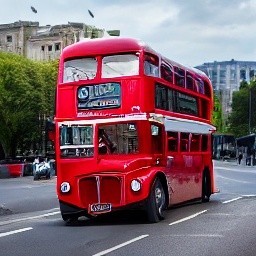}  \\
\includegraphics[height=0.3\linewidth, width=0.3\linewidth]{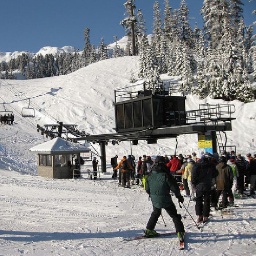}  &
\includegraphics[height=0.3\linewidth, width=0.3\linewidth]{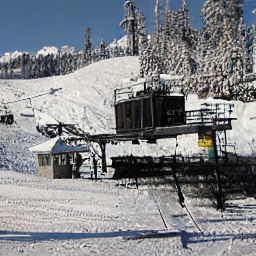} &
\includegraphics[height=0.3\linewidth, width=0.3\linewidth]{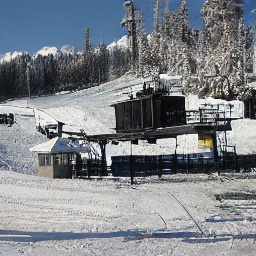} \\
\multicolumn{3}{c}{\centering Remove all persons} \\
\includegraphics[height=0.3\linewidth, width=0.3\linewidth]{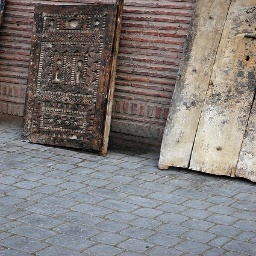}  &
\includegraphics[height=0.3\linewidth, width=0.3\linewidth]{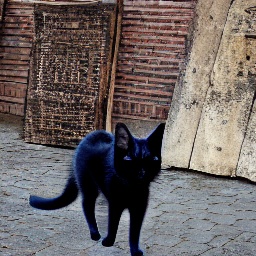} &
\includegraphics[height=0.3\linewidth, width=0.3\linewidth]{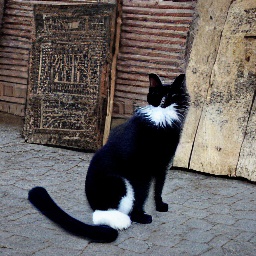} \\
\multicolumn{3}{c}{\centering Add a black and white cat}\\
\end{tabular}
}
\endgroup
\caption{\textbf{Qualitative Comparison of ControlNet-xs and ControlNet-Vxs.}}
\label{fig:qual-res-ctrl}
\vspace{-5mm}
\end{figure}

\subsection{Comparison of ControlNet-xs Variants}
\label{res:rq3}
\begin{table}[!h]
\centering
\resizebox{0.91\linewidth}{!}{
\begin{tabular}{lllll}
\hline
               & CLIP-T $\uparrow$ & CLIP-Aes $\uparrow$ & LPIPS $\downarrow$ & FID $\downarrow$    \\ \hline
ControlNet-xs  & 0.2639 & 4.9824   &  0.4588     & 17.143 \\
ControlNet-Vxs & \textbf{0.2648} & \textbf{5.0516}   & \textbf{0.4362}      & \textbf{15.125} \\ \hline
\end{tabular}
}
\caption{\textbf{Canny-to-Image Results.} Results are reported on the COCO validation set. We use the best performing configuration of add-concat based fusion for ControlNet-xs.} 
\label{tab:s2i}
\vspace{-3mm}
\end{table}

In this section, we investigate the impact of incorporating Volterra fusion into ControlNet-xs. Our assessment involves validating this enhancement through performance evaluation on two distinct tasks: i) Instruction-Based Image Editing and ii) Canny-to-Image Generation. The results presented in Table \ref{tab:cnet-gedit} demonstrate that our proposed variant, ControlNet-Vxs, yields significant performance improvements, particularly in the remove task (a reduction of $11.4 \%$ in FID score), while also enhancing metrics in the object addition task. These findings are further corroborated in the Canny-to-Image generation task (Table \ref{tab:s2i}). The notable enhancement observed in add-remove tasks underscores the efficacy of our approach in handling more complex non-linear tasks.
\section{Conclusion and Future Work}
In this work, we propose GalaxyEdit, a large scale dataset for object addition and removal tasks. A diffusion model trained on this dataset surpasses the performance of leading methods for add and remove tasks. Furthermore, we introduce ControlNet-Vxs - a novel Volterra fusion based ControlNet-xs. The introduction of non-linear Volterra filters allows the model to adapt to more complex conditioning, such as natural image control inputs. Through extensive experiments on multiple datasets we establish the efficacy of our proposed model. Despite our method's strong qualitative and quantitative results, it has some  limitations. The accuracy of our pipeline is constrained by individual module capabilities. While GalaxyEdit originates from the COCO dataset, our data generation pipeline is adaptable to other image datasets. Future extensions could encompass advanced 3D spatial edits necessitating robust spatial reasoning abilities. The proposed adapter variant shows promise for application in various tasks benefiting from non-linear network design. We aim for our work to inspire future research endeavors addressing the challenges and opportunities highlighted in our study.
{
    \small
    \bibliographystyle{ieeenat_fullname}
    \bibliography{main}
}


\end{document}